# ADCNet: a unified framework for predicting the activity of antibody-drug conjugates

Liye Chen[1,2], Biaoshun Li[1,2], Yihao Chen[1,2], Mujie Lin[1], Shipeng Zhang[1], Chenxin Li[1], Yu Pang[1] and Ling Wang[1,*]

[1]*Joint International Research Laboratory of Synthetic Biology and Medicine, Ministry of Education, Guangdong Provincial Key Laboratory of Fermentation and Enzyme Engineering, Guangdong Provincial Engineering and Technology Research Center of Biopharmaceuticals, School of Biology and Biological Engineering, South China University of Technology, Guangzhou 510006, China.*

*Corresponding author: Ling Wang, E-mail: lingwang@scut.edu.cn.

[2]These authors contributed equally.

**Abstract**： Antibody-drug conjugate (ADC) has revolutionized the field of cancer treatment in the era of precision medicine due to their ability to precisely target cancer cells and release highly effective drug. Nevertheless, the realization of rational design of ADC is very difficult because the relationship between their structures and activities is difficult to understand. In the present study, we introduce a unified deep learning framework called ADCNet to help design potential ADCs. The ADCNet highly integrates the protein representation learning language model ESM-2 and small-molecule representation learning language model FG-BERT models to achieve activity prediction through learning meaningful features from antigen and antibody protein sequences of ADC, SMILES strings of linker and payload, and drug-antibody ratio (DAR) value. Based on a carefully designed and manually tailored ADC data set, extensive evaluation results reveal that ADCNet performs best on the test set compared to baseline machine learning models across all evaluation metrics. For example, it achieves an average prediction accuracy of 87.12%, a balanced accuracy of 0.8689, and an area under receiver operating characteristic curve of 0.9293 on the test set. In addition, cross-validation, ablation experiments, and external independent testing results further prove the stability, advancement, and robustness of the ADCNet architecture. For the convenience of the community, we develop the first online platform (https://ADCNet.idruglab.cn) for the prediction of ADCs activity based on the optimal ADCNet model, and the source code is publicly available at https://github.com/idrugLab/ADCNet.

## 1. Introduction

With the in-depth study of molecular biology, target-oriented drug discovery has become mainstream, with the aim of discovering more selective and promising clinical candidates[1]. Among them, the traditional small molecule inhibitor-targeted therapeutic strategy has achieved remarkable results[2], whereas there are still some fatal shortcomings, including off-target effects[3], limited therapeutic windows[4], and drug resistance concerns[5,6]. These drawbacks underscore the need for innovative solutions in the field of targeted therapies. Remarkably, the specific expression of antigens has dramatically facilitated the development of monoclonal antibodies[7], partially overcoming the limitations of traditional small-molecule inhibitors[8]. Despite the notable advances made by monoclonal antibodies in cancer treatment, their utility against highly heterogeneous cancer cell populations remains elusive[4].

In response to such a challenge, antibody-drug conjugate (ADC) has emerged as an innovative drug design concept [9], that ingeniously combines the highly targeted nature of monoclonal antibody with the cytotoxicity of toxin by means of suitable linker[10]. Such precise anchoring and direct cytotoxicity dramatically increase the therapeutic effect on cancer while reducing the damage to normal cells. Typically, the antibody portion of an ADC provides a unique targeting for cells carrying the desired antigen. When the ADC binds to the surface of the intended cell, the cell initiates endocytosis and transports the ADC to the lysosome. Subsequently, the cytotoxic portion of the ADC is released into the cell to exert its cytotoxic effects[11]. Tracing the evolution of ADCs[12], vindesine-anti-CEA conjugate was first applied in human clinical trials as early as 1983[13]. As the field continues to advance, the design of ADCs has been progressively refined over several generations. First-generation ADCs, such as gemtuzumab ozogamicin, typically contain a humanized antibody, a single toxin structure, and acid-unstable linker[14], where the humanized antibody and toxin evolved from murine-derived antibodies and conventional chemotherapeutic agents, respectively[15]. Their conjugation patterns are commonly based on random conjugation of lysine and cysteine, producing mixtures with uneven drug-antibody ratio (DAR), which results in poorly defined therapeutic index criteria[16]. Obviously, first-generation ADCs are prone to unmanageable off-target toxicity and narrow therapeutic windows due to their characteristics. In contrast, second-generation ADCs represented by brentuximab vedotin and trastuzumab emtansine have introduced more potent cytotoxicity payloads like tubulin polymerization inhibitors[17], as well as steadier linkers[18], guaranteeing both clinical efficacy and stability to a large extent. In this context, the remarkable breakthrough of third-generation ADCs, such as polatuzumab vedotin and enfortumab vedotin, is precisely the application of site-specific coupling technology[19] and the adoption of hydrophilic linkers[20], which not only make the DAR controllable, but also greatly enhance the safety and anticancer activity[18]. Nevertheless, opportunities and challenges normally accompany each other in this filed. Constraints such as the choice of antibodies, toxic payloads, and the design of linkers have contributed substantially to the difficulties in the rational design of ADCs and the continued expansion of therapeutic index[21-23].

In the rational design of ADCs, it is particularly important to explore the relationship between the conformations and properties of ADCs through computational methods such as homology modelling, molecular docking, and molecular dynamics simulations, providing new perspectives for further accelerating the research process of ADCs and understanding the mechanism of action (MOA)[24]. For example, Nimish et al. proposed the use of a combination of molecular docking and molecular dynamics simulations to design a non-covalently bound MAGNET linker platform that largely preserves the ability of ADCs to bind to antigens and overcomes the limitations of existing synthetic strategies[25]. Molecular docking is also used to simulate the binding of a drug-linker to an antibody to determine the optimal site for conjugation, and the antibody could then be engineered to introduce cysteine residues at the immobilization site for more purposeful site-specific coupling[26]. Computational methods have been also used to investigate the MOAs of existing ADCs. Ruby et al. explored the MOAs of nine Food and Drug Administration (FDA)-approved ADCs for cancer therapy by integrating multiple computational approaches, including ADMET risk prediction, potential target identification, molecular docking, molecular dynamics simulations, as well as non-covalent interactions in ADCs[27]. Ultimately, they highlighted the high toxicity of the payloads in these ADCs and identified specific binding sites on the target antigens. In addition, a traditional machine learning (ML) method has been used to mine human gene expression data to optimize the rational selection of breast cancer ADC targets[28]. Although current computational methods provide valuable assistance in designing ADC drugs and/or elucidating their mechanisms, they are often tailored to specific ADCs, restricting their applicability in general. Recently, deep learning (DL) methods are widely and successfully applied in the fields of prediction and exploration of structure-activity/property relationships[29-33], molecular property prediction[34-41], as well as ADMET prediction[42].

Currently, to the best of our knowledge, the application of DL methods in the field of ADCs has not been reported due to the lack of standardized and unified data in the ADCs field. Encouragingly, with the accumulation of ADC research data, Zhu et al. recently presented the first comprehensive online database called ADCDB to store ADCs and associated data[43]. To date, the ADCDB contains 6,514 ADCs, 1,157 antibodies, 493 linkers, 446 cytotoxic payloads, as well as their corresponding pharmacology assay data. Obviously, the database provides a data basis for developing computational tools for the rational design of ADCs.

In the present study, we propose a unified and generalized deep neural network model known as ADCNet for activity prediction of ADCs. As shown in Fig. 1, it is capable of efficiently predicting the anticancer cellular activity of targeted ADCs based on protein sequence representations of antigens and antibodies, SMILES representations of linkers and payloads, together with DAR values that are highly correlated with ADCs activities. To comprehensively learn and understand the interrelationships between the different components, ADCNet inputs different parts of a given antigen-ADC complex into the embedding layer separately to obtain distributed representations, which are then embedded into separate neural network modules for feature processing. The feature spliced vectors are fed to a multilayer perceptron (MLP) with two fully connected layers (FCL) for processing to generate the final prediction. ADCNet performed excellently on the test set, with an average accuracy (ACC), area under the receiver operator characteristics curve (AUC), and a balanced accuracy as high as 87.12%, 0.9293 and 0.8689, respectively. We further validated the ADCNet model by using 19 newly reported ADCs that involved 5 antigens, 5 antibodies, 18 linkers, and 5 payloads. Among these 19 ADCs, the model can successfully predict the anticancer capacities of 18 ADCs, thus achieving a prediction accuracy of 94.74%. Finally, we provide a free online platform to support the rational design and activity prediction of new ADCs.

## Fig. 1: A diagram illustrating the ADCNet framework

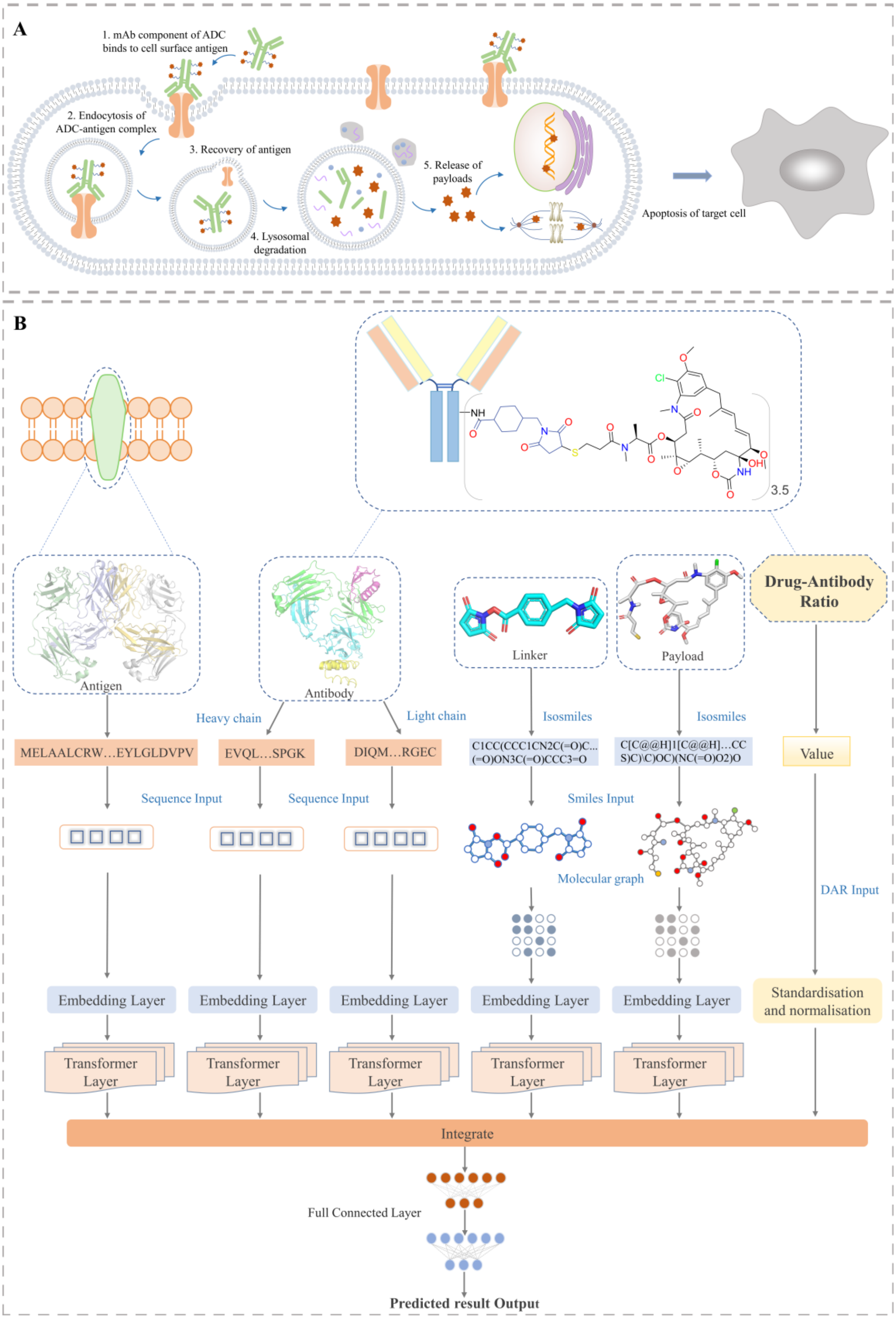

(**A**), The mechanism of action of ADCs. (**B**), The network architecture of ADCNet model.

## 2. Materials and methods

### 2.1 ADCs dataset preparation and labeling

The data set used to construct the ADCNet model was uniformly sourced from ADCDB[43]. The original data was processed with the following steps: 1) ADCs with complete structural information on each item were retained, ensuring that the four important structural information of each ADC were available, including the heavy and light chain sequences of the antibody, the antigen sequence, and the isomeric SMILES of the linker and payload, while ADCs with incomplete information on the above were directly discarded; 2) removing duplicates, cleaning and standardizing the raw data; 3) uniformly converting units of bioactivity testing data including $IC_{50}$, $EC_{50}$, and $GI_{50}$ (e.g., μM, pM, etc.) at cellular level into nM; 4) if an ADC has multiple biological activity data, removing invalid values such as ">" and "<" and keeping and selecting the minimum value as the final value. Subsequently, we labelled the data as follows 1) if an ADC is in research status of marketed, clinical phases I, II, and III, we all labelled as "positive"; 2) if a ADC is in the investigational stage, it was labelled as "positive" according to its $IC_{50}$, $EC_{50}$, and $GI_{50}$ <= 100 nM, and vice versa (labelled as "negative"). Finally, the obtained ADC dataset was randomly divided into three subsets: 80% for training, 10% for validation, and 10% for testing.

### 2.2 ADCNet architecture and training strategies

#### 2.2.1 Feature representations of antigen and antibody in ADC

In the current study, the antibodies in ADCs as well as the antigens they act on, were uniformly classified as proteins based on their structural characteristics. The protein language model ESM-2 (freely available at https://github.com/facebookresearch/esm), developed by the Facebook team, is an unsupervised and Transformer-based pre-trained model that enables the prediction of protein structure from protein amino acid sequences and has achieved state-of-the-art (SOTA) performance across multiple downstream task experiments[44]. The ESM-2 model was used to efficiently extract features of antibodies and their corresponding antigens in this study. Specifically, the model processed the input sequences through its multi-layer neural network architecture, that is, the features of amino acids were abstracted and processed layer by layer. Finally, ESM-2 extracted features of the corresponding protein sequences through transfer learning and outputs high-quality 1280-dimensional feature vectors, which can be effectively applied to the relevant representation of ADCs.

### 2.2.2 Feature representations of payload and linker in ADC

Recently, we introduced a generalized and self-supervised pretraining DL framework called functional group-based bidirectional encoder representations from transformers (FG-BERT) to learn meaningful representation of molecules from ~1.45 million unlabeled drug-like molecules[36]. The model showcases SOTA predictive capabilities across 44 benchmark datasets covering biological activities, physicochemical properties, physiology, and toxicity attributes. In this study, the payload and linker belonging to the ADC were rationally divided into small molecule items based on their structures. Accordingly, we adopted the FG-BERT DL model (freely available at https://github.com/idrugLab/FG-BERT) to deeply comprehend and predict the chemical properties and biological activities of payloads and linkers in ADCs. FG-BERT is remarkable for its intensive analysis of intramolecular functional groups, which is critical for the accurate assessment of the efficacy and stability of ADCs. Initially, the input SMILES string of each payload and linker in ADC is converted into molecular graph to capture molecular structural features as well as information about the connections between atoms. In this molecular graph, atoms are represented as nodes, while chemical bonds are considered as edges. Subsequently, the FG-BERT model was fine-tuned to learn ADC-related representations, and useful molecular representations were finally extracted through transfer learning. After this process, the small molecule structures of linkers and payloads are mapped into 256-dimensional eigenvectors, providing important molecular-level information for ADC studies.

2.2.3 Feature representation of DAR

DAR essentially represents the average number of drug molecules attached to a single antibody. Previous studies have pointed to the profound impact of this metric on multiple aspects of ADC efficacy, safety, and stability[45,46]. Therefore, we specifically added DAR as a key input variable to the model. To increase the accuracy and effectiveness of data processing, we used the StandardScaler tool to standardize the extracted DAR data columns, ensuring that each variable in the dataset has a uniform zero mean and unit variance, thus eliminating the possible bias of different magnitudes of data. In addition, to further enhance the usability and consistency of the data, the standardized data was normalized by the normalize function in the TensorFlow framework. While this step ensures that the data is normalized to a paradigm of 1 on the specified axes, the model is able to capture the characteristics of the data more accurately when analyzing and interpreting this data, ensuring the accuracy and reliability of the findings.

2.2.4 ADCNet framework

As shown in Fig. 1B, ADCNet is a multi-channel feature fusion DL model that integrates the representations of antigens, antibodies, linkers, payloads and DAR values. Remarkably, both ESM-2 and FG-BERT introduced in ADCNet are based on the BERT large language model, which are typically characterized by the inclusion of embeddings, Transformer encoders, and fully connected layers. The principles and specific implementations of the FG-BERT[36] and ESM-2[44] models have been specifically shown in previous studies. In brief, FG-BERT is pre-trained by masking-recovery of functional groups using a corpus of ~1.45 million unlabeled small molecules, while ESM-2 is pre-trained by masking-recovery of amino acids using a corpus of ~65 million unlabeled unique protein sequences. The two pre-trained models are then utilized to predict downstream tasks through transfer learning. In ADCNet, we use ESM-2 for feature extraction of large molecule antibodies and antigens. With this operation, we get the feature vectors $t_1$ (heavy chain), $t_2$ (light chain) for antibodies, and $t_3$ for antigens. Meanwhile, FG-BERT is used for feature extraction of small molecule linkers and payloads, and we then get the feature vector $x_1$ for linker and feature vector $x_2$ for payload. DAR is standardized and normalized to the -1–1 interval to get $t_4$. We then splice these feature vectors using the concat function, and the formula is as follows:

$$x = concat(x_1, x_2, t_1, t_2, t_3, t_4) \quad (1)$$

The final feature fusion is performed by Multilayer Perceptron (MLP) for prediction of ADC activity, and the formula is defined as follows:

$$\hat{y} = Wx + b \quad (2)$$

Where $\hat{y}$, $W$, $x$, $b$ represent the output vector, weight matrix, input vector and bias vector, respectively. In addition, LeakyReLU is used as an activation function in the ADCNet model. The detailed output dimensions of each layer of ADCNet are provided in Supplementary Table S1.

2.2.5 Model training protocol and hypeparameters optimization

The entire ADCNet model is implemented using Python and Tensorflow[47] with a unique training strategy. During model training, the dataset is divided into training, validation and test sets, which are split using three different random seeds with a ratio of 8:1:1. Once all data is loaded, the architecture of shared models and the separate output layers is defined. To avoid model overfitting during training, ADCNet was trained using an early stopping strategy, where the number of training epochs was set to 200 and the tolerance value was set to 30. Concretely, after each round of training, the AUC value of the validation set is evaluated and recorded. If the AUC value of the validation set fails to improve within 30 consecutive epochs, the training loop is immediately interrupted and the test set is entered for final testing. After each training iteration, we check whether the current validation set AUC value exceeds the previous best AUC value. If this value is exceeded, the current model weights are saved as the new best model. If it is not exceeded, the stopping monitor will increase by one. When the value of the tolerance detector reaches 30, the training process ends and the test set is evaluated using the saved best model weights. If the tolerance detector still does not reach 30 throughout 200 epochs, the training also ends and the performance evaluation of the test set is finally performed. This process aims to ensure that the model does not fit while learning the data in order to obtain the model with the best generalization performance on the test set. During the training process of ADCNet, Bayesian optimization technique based on Hyperopt and Python packages was introduced to improve and tune the hyperparameters of the model. The optimal hyperparameters were determined by performing 30 searches in the defined parameter space (Supplementary Table S2) to represent the best configuration of the model in terms of final performance results.

To comprehensively and impartially evaluate the predictive performance of ADCNet, we constructed four classical basic ML models for comparison, including logistic regression (LR)[48], random forest (RF)[49,50], support vector machine (SVM)[51], and extreme gradient boosting (XGBoost)[52]. The characteristics of both linker and payload in the ADC are represented by two molecular fingerprints, namely the 166-bit MACCS keys[53] and the 1024-bit Morgan fingerprint[54]. The RF-, SVM-, XGBoost-, and LR-based models were built on the basis of the fingerprints of linker and payload, protein features of antigen and antibody, and DAR value. The scikit-learn library in Python (freely available at https://github.com/scikit-learn/scikit-learn, version 0.24.1) was used to construct the RF, SVM, and LR models, while the XGBoost Python library (freely available at https://github.com/dmlc/xgboost, version 1.3.3) was utilized to build the XGBoost model. All these traditional ML and as the ADCNet DL models were trained on GPU [NVIDIA Corporation GV100GL (Tesla V100 PCIe 32 GB)] and CPU (Intel(R) Xeon(R) Silver 4216 CPU @ 2.10 GHz).

## 2.3 Evaluation of model performance

The performance of ADCNet and traditional baseline ML models is scored by a variety of metrics, including positive predictive value (PPV), negative predictive value (NPV), specificity (SP/TNR), sensitivity (SE/TPR/Recall), ACC, F1 score, BA, Matthews correlation coefficient (MCC), area under the precision-recall curve (PR-AUC), and AUC. These metrics are defined as follows:

$$PPV = \frac{TP}{TP + FP} \tag{3}$$

$$NPV = \frac{TN}{TN + FN} \quad (4)$$

$$SE = \frac{TP}{TP + FN} \quad (5)$$

$$SP = \frac{TN}{TP + FP} \quad (6)$$

$$ACC = \frac{TP + TN}{TP + TN + FP + FN} \quad (7)$$

$$F1 = \frac{2 \times TP}{2 \times TP + FN + FP} \quad (8)$$

$$BA = \frac{TPR + TNR}{2} = \frac{SE + SP}{2} \quad (9)$$

$$MCC = \frac{TP \times TN - FN \times FP}{\sqrt{(TP + FN) \times (TP + FP) \times (TN + FN) \times (TN + FP)}} \quad (10)$$

Where TP, TN, FP, FN, TNR, and TPR represent true positive, true negative, false positive, false negative, true negative rate, and true positive rate, respectively. Furthermore, AUC is the area under the ROC curve plotted by calculating TPR vs. FPR at different discrimination thresholds, which is used to measure the overall performance of the model.

### 2.4 Web server implementation

We have developed an online platform that facilitates users to predict ADC activity for free. The platform consists of three main parts: front-end, back-end, and cloud server. User input is received using Streamlit framework and passed to the back-end Tensorflow model for processing; and the cloud server runs on CentOS and uses Nginx to handle user requests. The platform is hosted on Tencent Cloud to ensure its high speed and stable performance.

## 3. Results and discussion

### 3.1 Analysis and visualization of ADC modelling dataset

As described in the Methods section, we have defined for the first time a standard for collecting and processing ADC modeling dataset, based on the fact that the efficacy of a given ADC is maintained by the complexities of how antibody, linker, and payload components interact with tumors and their microenvironment[10,55], all of which have significant clinical implications. As shown in Fig. 2A, a total of 435 ADCs were obtained based on the mentioned standard. Fig. 2B shows the research phase and corresponding proportions of these ADCs.

When constructing predictive models, the structural diversity and wide chemical space of the molecules in the dataset are crucial for improving the accuracy and robustness of the models[56]. Hence, based on the characteristics of the ADCNet architecture, we performed statistical analysis of the antibodies, linkers, payloads, DAR, and antigens involved in the ADCs. As shown in Fig. 2C, among these 435 ADCs, the number of unique antibodies, linkers, payloads, DAR values, antigens is 154, 82, 71, 75, and 66, respectively, indicating the overall diversity of ADC modelling dataset used in this study. Fig. 2D exhibits that 66 antigens can be divided into six different categories based on sequence alignment results, which intuitively understands and demonstrates the structural and functional diversity of these antigens in the construction of the ADCNet model. The top five antigens are ERBB2 (106), KIT (48), FGFR2 (36), TNFRSF1A (33), and EGFR (22). Meanwhile, humanized IgG1-kappa is most widely used to design ADCs (Fig. 2E), followed by Chimeric IgG1-kappa, humanized IgG1, and humanized IgG1-nd. Supplementary Fig. S1 displays the distribution of isoelectric point (pI) values and molecular weight (MW) of antigens and antibodies, further indicating the diversity of antibodies and antigens.

We further performed structural diversity and chemical space analysis on 82 linkers and 71 payloads. As shown in Fig. 2F, Bemis–Murcko scaffold analysis[57] shows that the proportion of the scaffolds for linkers and payloads in the ADC dataset is between 58.54% and 67.61%, demonstrating that the linkers and payloads are structurally more diverse. In addition, the chemical space of the linkers and payloads in the ADC dataset can be characterized by using MW and AlogP. Fig. 2G and 2H show that both linkers and payloads are distributed over a wide range of MW (linkers: 120.195–1702.11, payloads: 233.097–1580.593) and AlogP (linkers: -6.504–6.908, payloads: -4.675–7.406), indicating that the linkers and payloads in the ADC dataset have a broad chemical space. Collectively, the above analysis results, together with 75 different DAR values (Fig. 2C), demonstrated that the five input components in the ADCNet framework are also diverse.

Among these ADCs, 281 ADCs were labeled as positives and 154 ADCs were labeled as negatives. Finally, the whole ADC modelling dataset was randomly split into the training (80%), validation (10%) and test (10%) sets. The training set is used to build a predictive model, the validation set is used to optimize and select hyperparameters, and the test set is utilized to test the final predicted performance of the model.

**Fig. 2: The overall analysis results of the ADC modelling dataset.**

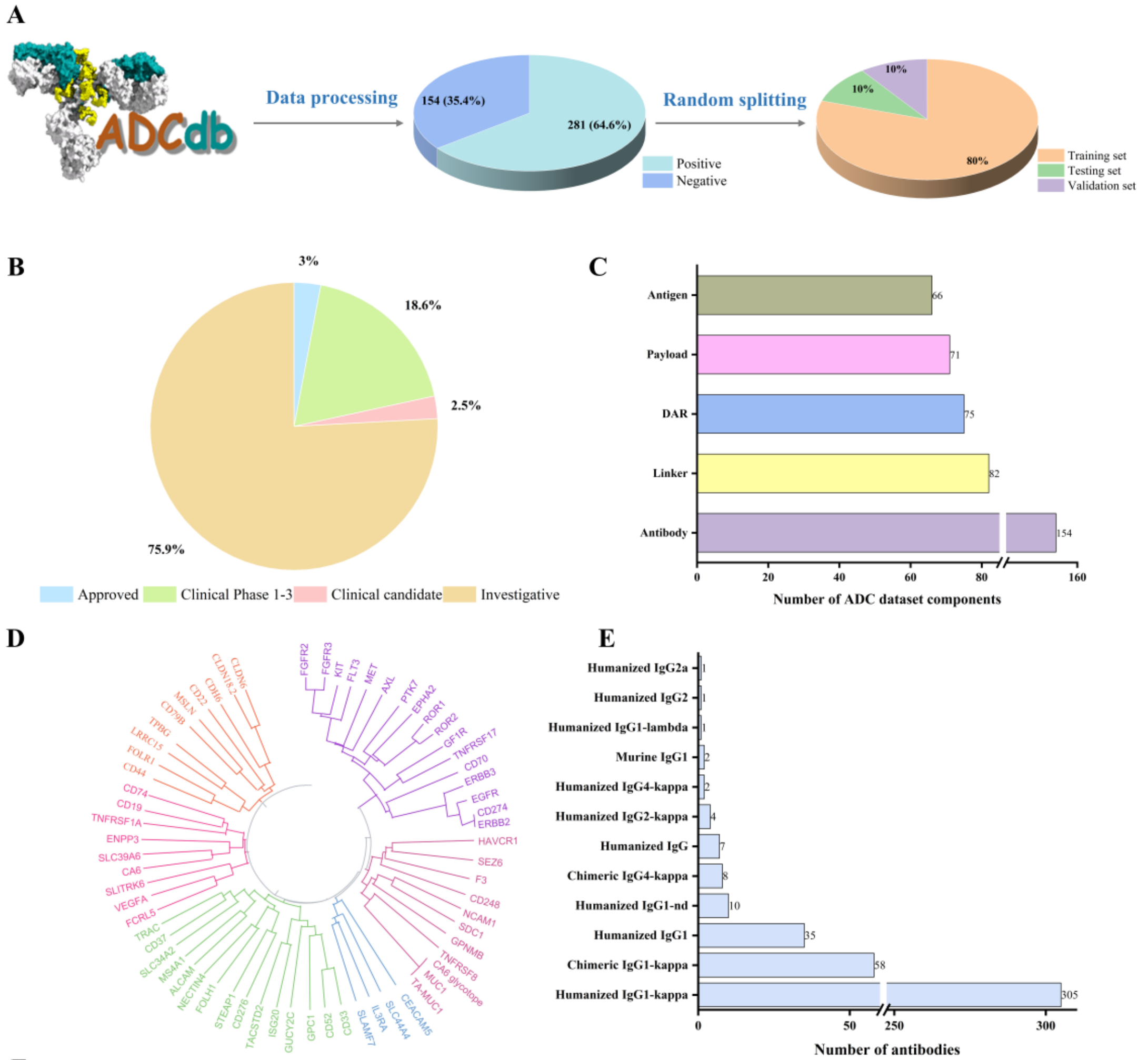

F

| | No. of compounds | No. of scaffolds | Scaffolds/compounds (%) |
|---|---|---|---|
| Payloads | 71 | 48 | 67.61 |
| Linkers | 82 | 48 | 58.54 |

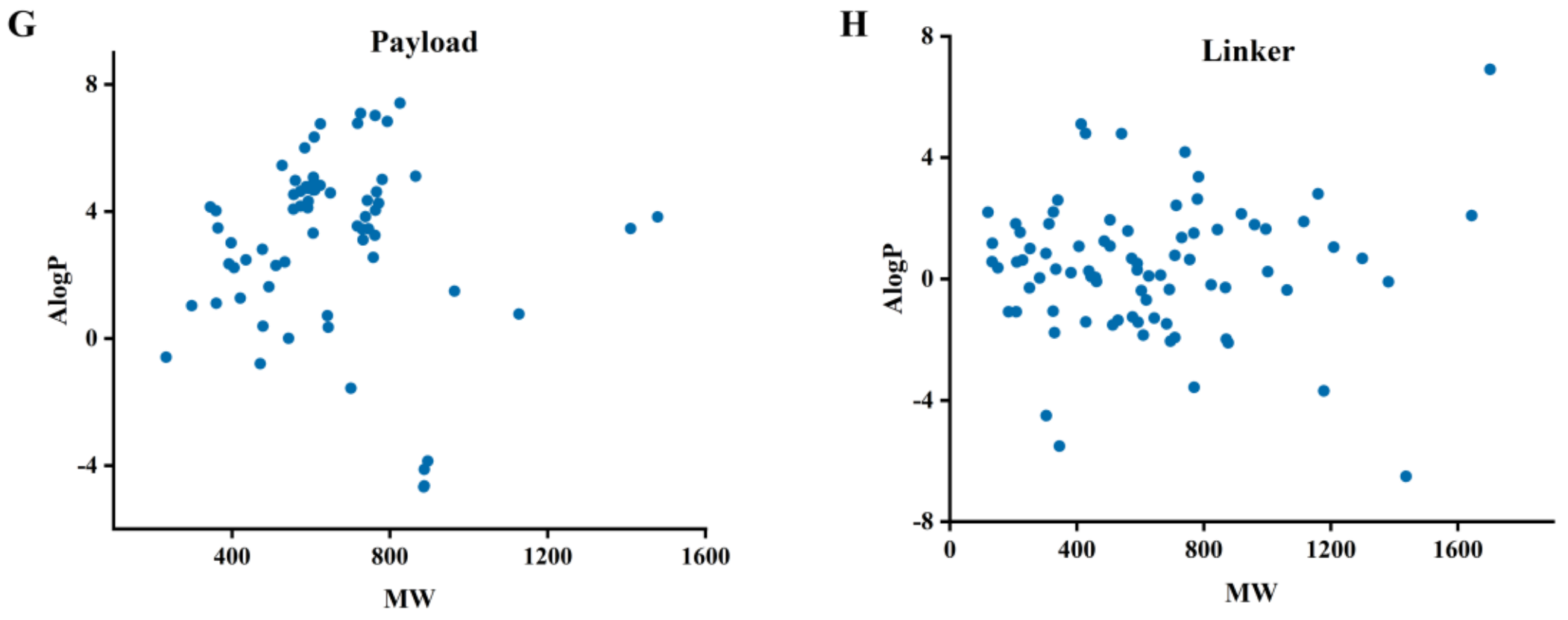

(**A**), Data acquisition and processing procedure. (**B)**, Research distribution of ADCs in the modelling dataset. (**C**), The number of components involved in antigens, antibodies, linkers, payloads, and DAR values in the ADC dataset. (**D**), Developmental tree analysis of antigens in the ADC dataset. (**E**), The number of each antibody class in the ADC dataset. (**F**), Scaffold diversity analysis of payloads and linkers in the ADC dataset. Chemical space analysis of payload (G) and linker (H) in the ADC dataset. The developmental tree was generated by MEGA11 software (https://www.megasoftware.net/). The scaffold of each payload or linker was calculated by RDKit software (https://www.rdkit.org/). The chemical space was defined using molecular weight (MW, X-axis) and AlogP (Y-axis). MW and AlogP were computed by RDKit software (https://www.rdkit.org/).

## 3.2 Performance of the ADCNet model

We utilized the carefully designed ADC dataset to evaluate the predictive ability of the ADCNet model. Detailed performance results of the ADCNet and eight classic ML models are provided in Table 1. Notably, all models are constructed based on the same dataset, data split method and initially seeds for fairly comparison. As shown in Table 1, the ADCNet achieves the best performance on the test set due to its highest values on all commonly used evaluation metrics. For one thing, the ADCNet shows the best overall accuracy compared to the baseline models, with the highest ACC, AUC, and PR-AUC values of 0.8712, 0.9293, and 0.9518, respectively. In addition, the ADCNet achieves the maximum values of SE (0.8713), SP (0.8665), PPV (0.9241), and NPV (0.7778), indicating that the prediction accuracy of the model for both ADC positives/actives and negatives/inactives is not only optimal, but also balanced, compared with the traditional ML models.

For another, the results of data collection and analysis have shown that the current ADC dataset is unbalanced, due to its inclusion of 281 actives and 154 inactives. Therefore, MCC, F1 and BA indicators were employed to further evaluate the ADCNet model. Compared to eight traditional ML-based baseline models, the ADCNet model performed the best, with the highest values of MCC (0.7196), F1 (0.8968), and BA (0.8689).

In fact, there are two main differences between the ADCNet model and the ML-

based baseline models. First, the baseline models use the Morgan fingerprint or MACCS fingerprint to encode the linkers and payloads in the ADCs, while the ADCNet model uses the FG-BERT large language model to learn the representation of the linkers and payloads. Previous study has shown the superiority of FG-BERT in molecular learning representation[36], which may be one of the reasons for the explosive performance of the ADCNet model. Second, in view of the complex input features from the five components (antigen, antibody, linker, payload, and DAR), the traditional ML methods can only accept all the input features, and then directly build the predictor and output the result, which obviously lacks the process of meaningful feature processing. In contrast, ADCNet as a DL model can learn meaningful features from the complex input features of these five components through self-learning and feedback mechanisms, which ultimately leads to the best performance in the ADCs activity prediction task. Incidentally, Supplementary Fig. S2 shows the iterative change in loss during the training of the ADCNet model and the change in AUC curves for the training and validation set.

**Table 1 Performance evaluation results of ADCNet and traditional ML-based models on the test set.**

| Model | SE | SP | MCC | ACC | AUC | F1 | BA | PRAUC | PPV | NPV |
|---|---|---|---|---|---|---|---|---|---|---|
| LR-MACCS | 0.8296 ± 0.0641 | 0.5318 ± 0.061 | 0.3791± 0.0303 | 0.7348 ± 0.0347 | 0.6807 ± 0.006 | 0.8068 ± 0.0387 | 0.6807 ± 0.006 | 0.8647 ± 0.0229 | 0.7862 ± 0.0239 | 0.6127 ± 0.0489 |
| LR-Morgan | 0.8527 ± 0.0442 | 0.6486 ± 0.0364 | 0.5105 ± 0.0352 | 0.7879 ± 0.0262 | 0.7506 ± 0.0111 | 0.8433 ± 0.0281 | 0.7506 ± 0.0111 | 0.8928 ± 0.016 | 0.8344 ± 0.0136 | 0.6859 ± 0.0422 |
| RF-MACCS | 0.8073 ± 0.0497 | 0.7028 ± 0.0728 | 0.4979 ± 0.1128 | 0.7727 ± 0.0601 | 0.7551 ± 0.0604 | 0.8259 ± 0.0548 | 0.7551 ± 0.0604 | 0.8908 ± 0.0422 | 0.8455 ± 0.0611 | 0.6405 ± 0.0453 |
| RF-Morgan | 0.8196 ± 0.042 | 0.7237 ± 0.0417 | 0.5315 ± 0.0581 | 0.7879 ± 0.0347 | 0.7716 ± 0.0304 | 0.838 ± 0.0343 | 0.7716 ± 0.0304 | 0.8995 ± 0.027 | 0.8581 ± 0.0397 | 0.6621 ± 0.0465 |
| SVM-MACCS | 0.7863 ± 0.0199 | 0.6218 ± 0.1251 | 0.4013 ± 0.1291 | 0.7348 ± 0.0473 | 0.7040 ± 0.0675 | 0.8004 ± 0.0293 | 0.7040 ± 0.0675 | 0.8728 ± 0.0231 | 0.8154 ± 0.0427 | 0.5794 ± 0.0836 |
| SVM-Morgan | 0.7859 ± 0.0527 | 0.6742 ± 0.0273 | 0.4492 ± 0.0817 | 0.7500 ± 0.0455 | 0.7301 ± 0.0391 | 0.8081 ± 0.0409 | 0.7301 ± 0.0391 | 0.8810 ± 0.0248 | 0.8321 ± 0.0299 | 0.6064 ± 0.0667 |
| XGB-MACCS | 0.8649 ± 0.0071 | 0.7237 ± 0.0417 | 0.5869 ± 0.0376 | 0.8182 ± 0.0227 | 0.7943 ± 0.0241 | 0.8647 ± 0.0216 | 0.7943 ± 0.0241 | 0.9103 ± 0.0216 | 0.8648 ± 0.0362 | 0.7206 ± 0.011 |
| XGB-Morgan | 0.8526 ± 0.0748 | 0.7445 ± 0.0279 | 0.5955 ± 0.0771 | 0.8182 ± 0.0455 | 0.7985 ± 0.0284 | 0.8616 ± 0.0444 | 0.7985 ± 0.0284 | 0.9118 ± 0.024 | 0.8726 ± 0.0253 | 0.7222 ± 0.0962 |
| ADCNet | **0.8713 ± 0.0660** | **0.8665 ± 0.0387** | **0.7196 ± 0.0974** | **0.8712 ± 0.0572** | **0.9293 ± 0.0464** | **0.8968 ± 0.0569** | **0.8689 ± 0.0522** | **0.9518 ± 0.0366** | **0.9241± 0.0470** | **0.7778 ± 0.0481** |

The best performing model for each metric is highlighted in bold.

## 3.3 Multi-level model validation

### 3.3.1 Cross-validation

It is common practice to employ cross-validation approach in the model training and evaluation process to ensure the robustness and general applicability of the model. Accordingly, we adopted a 5-fold cross-validation approach to further evaluate the predictive performance of the ADCNet model in this study, which is particularly well-suited for smaller dataset because it can make more efficient use of limited data resource. As shown in Fig. 3A, the average BA, AUC, and ACC of 5-fold cross-validation is 0.8477 ± 0.0078, 0.9281 ± 0.0076, and 0.8521 ± 0.015, respectively, which were consistent with the results obtained by the conventionally trained ADCNet model. In addition, other evaluation indicators also show a similar trend. All of these results demonstrate the robustness of the ADCNet model.

### 3.3.2 Activity threshold-based model performance analysis

In the present study, the ADCNet model was originally trained using a default active/inactive cutoff value of 100 nM for $IC_{50}$, $EC_{50}$, and $GI_{50}$. In order to further reduce the influence of artificial selection on the experimental results and to enhance the accuracy and reliability of the study, we introduced another active/inactive cut-off criterion (1000 nM). In other words, an ADC is considered active only if its $IC_{50}$, $EC_{50}$, or $GI_{50}$ satisfy the cut-off condition, or if it is in clinical studies and approved. We then retrained the ADCNet model based on this new truncation criterion. As shown in Fig. 3B, the retrained model based on the new criterion archives the average BA, AUC, PRAUC, and ACC values of 0.8812, 0.9301, 0.9552, and 0.9015, respectively, which were extremely close to the values obtained by using the original criterion. Other evaluation metrics also exhibit a similar trends. These findings confirm that our model exhibits good stability in response to different truncation criteria. Notably, for the sake of simplicity and the fact that ADC acts as a magic bullet due to its highly active nature, only the initial set of labels was elected to train the final ADCNet model.

**Fig. 3: Results of ADCNet model validation**

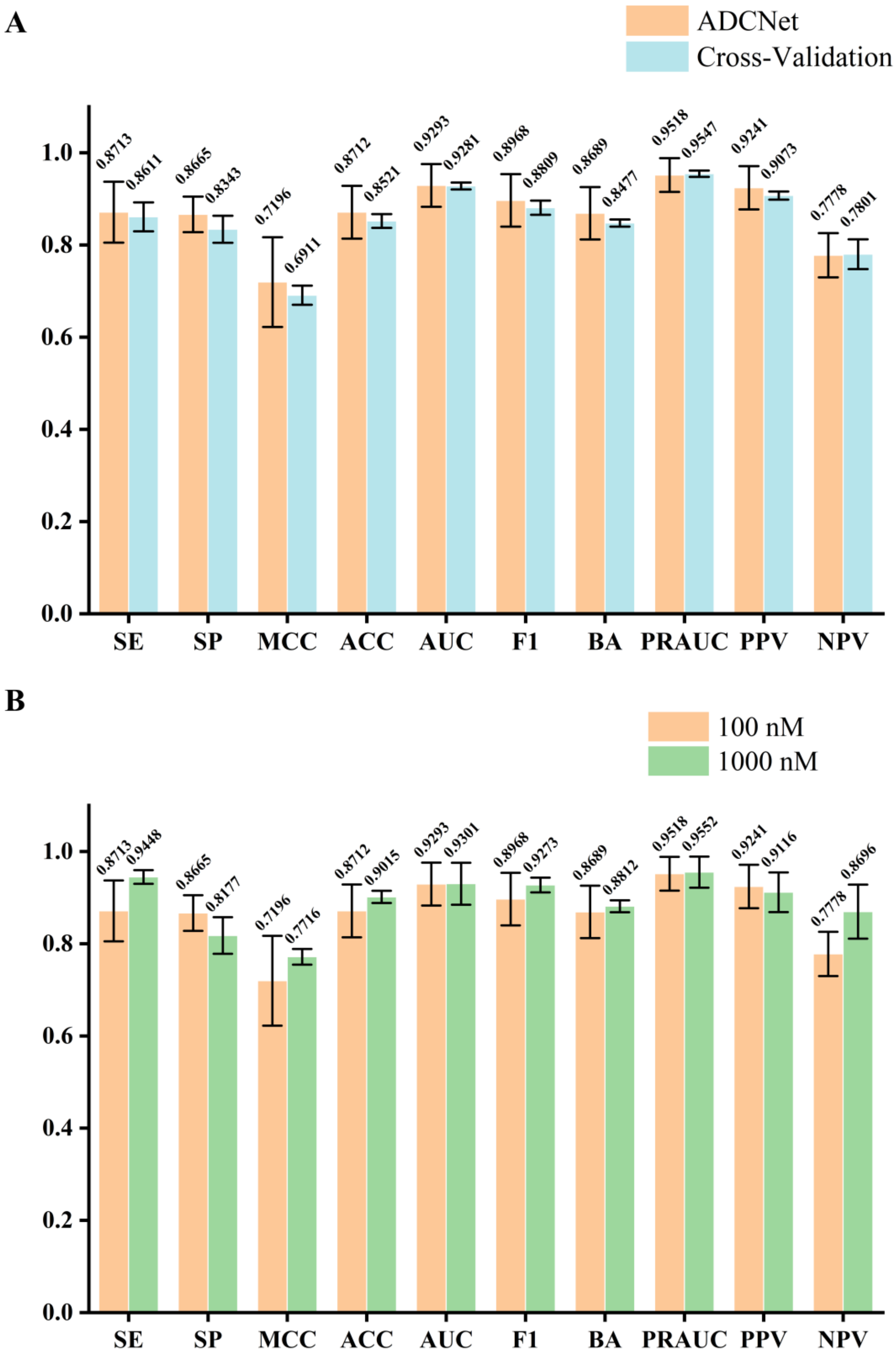

Comparison results of the original ADCNet model with the 5-fold cross-validation model (**A**) and the model based on a threshold of 1000 nM (**B**). The average metrics of three independent experiments on the test set were used to evaluate all models.

### 3.3.3 Model architecture validation based on ablation experiments

As shown in Fig. 1B, ADCNet is a multipath framework that combines information of the antibodies, linkers, payloads, DAR values, and antigens involved in the ADCs. To further investigate the effectiveness and necessity of these five well-designed modules in the ADCNet architecture, we perform a series of ablation studies to decouple ADCNet networks. Concretely, we designed five variants of ADCNet as follows:

1. ADCNet *without antigen information* (w/o antigen): removing the antigen module in Fig. 1B.
2. ADCNet *without antibody information* (w/o antibody): removing the antibody module in Fig. 1B.
3. ADCNet *without linker information* (w/o linker): removing the linker module in Fig. 1B.
4. ADCNet *without payload information* (w/o payload): removing the payload module in Fig. 1B.
5. ADCNet *without DAR information* (w/o DAR): removing the DAR module in Fig. 1B.

All these five variants of ADCNet were evaluated on the ADCNet dataset, and the experimental setup is consistent with the ADCNet for a fair comparison. Two key indicators, MCC and BA, were unitized to adjudicate the comparison of the final ablation experimental results due to the imbalance in the ADC dataset. As shown in Fig. 4A and 4B, the ADCNet model performed best according to the highest average MCC (0.7196) and BA (0.8689) values compared to the other five variants, demonstrating that the five well-designed modules are necessary in the ADCNet architecture and can complement each other to improve the final performance of the ADCNet model. For example, antigens are not intrinsic components of ADC molecules in principle, but the lack of antigen information module will cause the MCC and BA values of the variant model (w/o antigen) to decrease to 0.5155 and 0.738, with an overall decrease of 28.36% and 15.07%, respectively, compared to the complete ADCNet model. One possible explanation is that antigens play an important role in recognition of ADCs and guiding them into tumor cells, thereby supporting their anti-tumor activity. A similar trend was also observed on other variant models (Fig. 4A and 4B), verifying that the entire ADCNet architecture strongly dominates the simpler variants.

In addition, we used ACC and AUC indicators to judge the overall superiority of the ADCNet framework in ablation experiments. As shown in Fig. 4C, even if we do not consider the imbalance of ADC data, the ADCNet model also achieves the best ACC (0.8712) compared to the five variant models. Meanwhile, its AUC (0.9293) is also very close to the best AUC (0.9304) from the variant model (w/o DAR), with little difference (Fig. 4D). Collectively, these results further confirm the overall superiority and rationality of the ADCNet architecture proposed in this study.

**Fig. 4: Results of ablation experiments.**

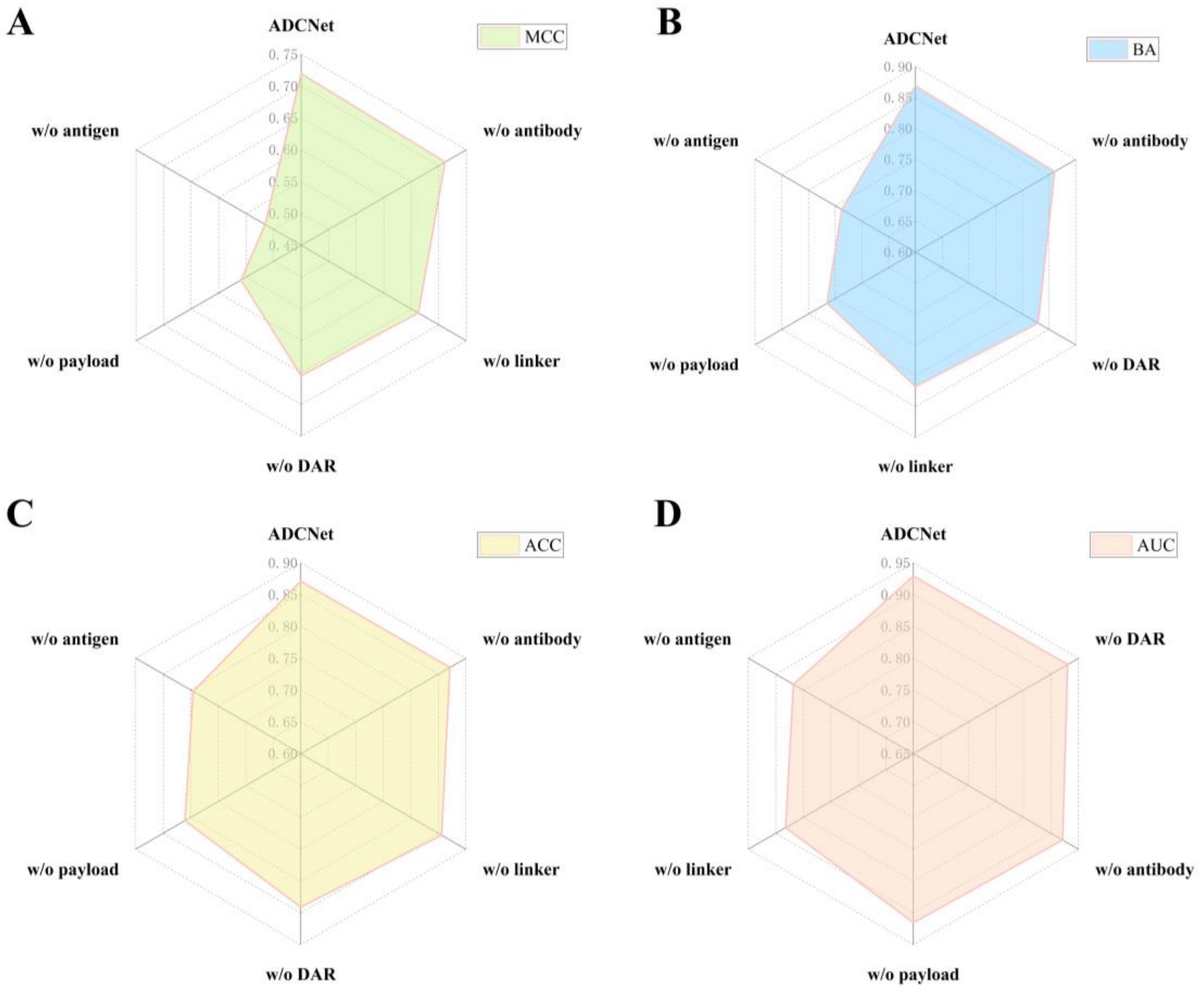

(**A**), (**B**), (**C**) and (**D**) represent the average MCC, BA, ACC and AUC values of the test set, respectively.

The ADCNet framework uses two pre-trained models to generate embeddings as input vectors for the prediction subnetworks. To evaluate its feature learning capability, we employed t-distributed stochastic neighbor embedding (t-SNE), an excellent dimensionality reduction method, to visualize a quintuple consisting of all embeddings of antibody, antigen, linker, payload, and DAR value. We applied the entire collected data as input for dimensionality reduction, and used the ADCNet model to reduce the embedding space of the ADCs in the before and after training scenarios. As shown Fig. 5B, the distribution of positive (green) and negative (orange) quintuples in the dataset shows that it is easy to separate positive and negative ADC samples by the ADCNet model, compared to the corresponding quintuple distribution without model training. Taken ADC1 (negative) and ADC2 (positive) as an example, although they have different activities and are structurally different (Fig. 5A), they overlapped in an untrained space. However, it was easy to separate and successfully classify in the space after the ADCNet training. Meanwhile, a pair of ADC molecules (ADC3 and ADC4, Fig. 5C) with opposite activities were randomly distributed before training, but both were successfully categorized after training. The more clearly the quintet of positive and negative roles can be separated, the better the categorization will be. Collectively, the embeddings of both categories are more prominent in terms of data distribution after training compared to before, which further verifies the excellent performance of our model.

**Fig. 5: Visualization of the ADC dataset using t-SNE method.**

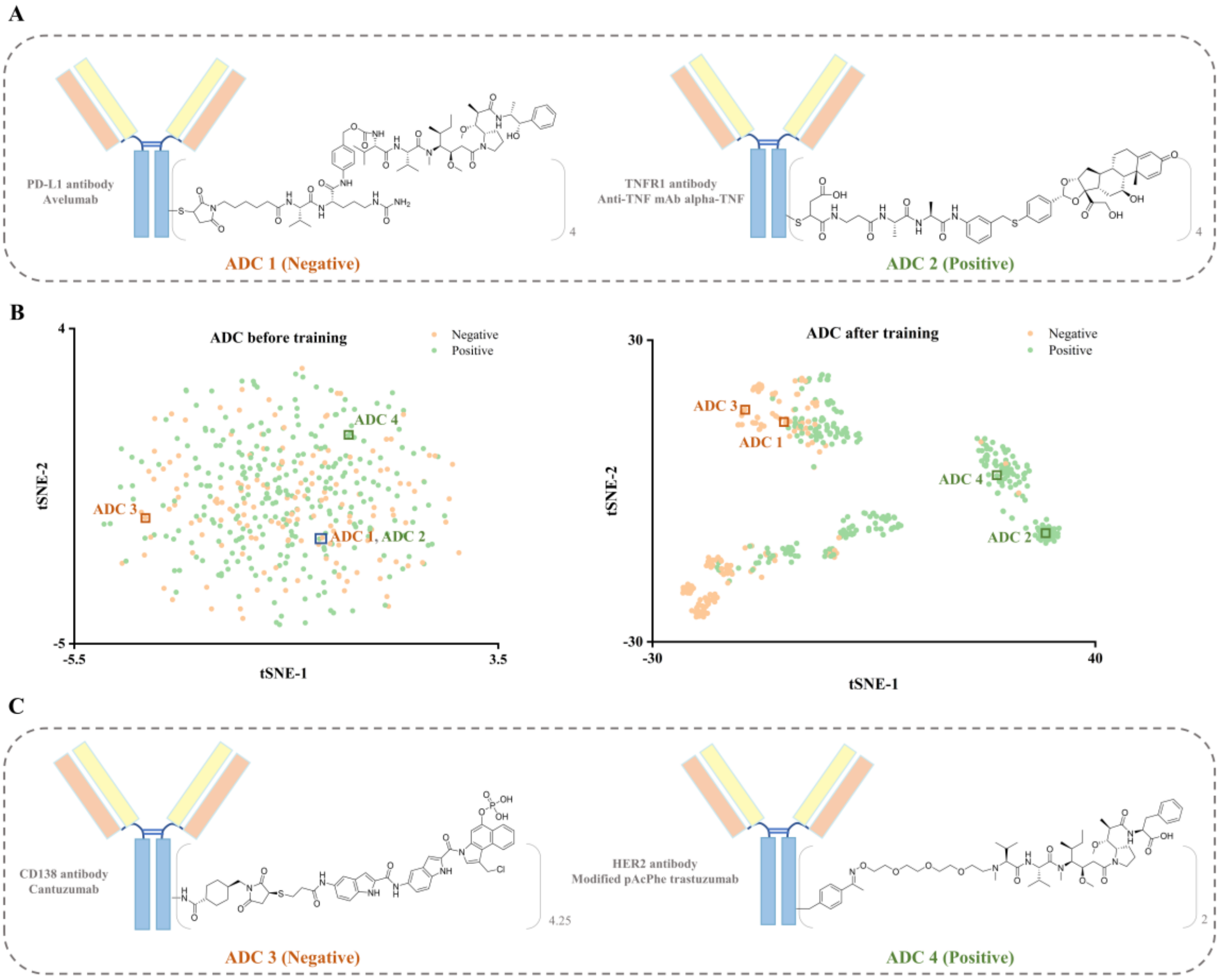

(**A**) represents ADC1 (PD-L1-ADC-2, negative) and ADC2 (Anti-TNF ADC 115, positive) with overlap in an untrained space. (**B**), Spatial distribution of the ADC dataset before and after training. (**C**) represent ADC3 (HuC242-SMCC-DC4, negative) and ADC4 (ARX-788, positive) with randomly distribution in an untrained space.

3.3.4 Independent testing based on new external dataset

The above results have been confirmed the robustness and stability of the ADCNet model, here, we continue to test the generalization ability of the ADCNet model because it is of great value for the application of the model. To this end, we collected an additional dataset of 19 ADCs that satisfy the defined criteria from the latest publications[25,58,59] and a patent (WO2023237050). Details of these 19 ADCs are provided in Supplementary Table S3.

As shown in Table 2, there are five antigens, five antibodies, 18 linkers, five payloads (Supplementary Fig. S3), and 16 DAR values in these ADCs. Subsequently, sequence alignment-based similarity score was used to assess the similarity of the antibody and antigen in each newly collected ADC to the corresponding items in the original ADC dataset, Meanwhile, ECFP-4 fingerprints-based molecular similarity calculation was conducted to evaluate the novelty of linker and payload in these 19 ADCs. Detailed results are listed in Supplementary Table S3. Table S3 shows that these new ADCs are derived from the redesign of specific items, including the substitution, modification, and/or recombination of antibodies, antigens, and payloads. However, our analysis also shows that the ADCs in the external dataset exhibit significant novelty in the overall structure, especially in terms of the structure of the linkers. In addition, to further assess the structural novelty and diversity of these 19 ADCs, the reconciled mean similarity as an evaluation metric was adopted. This similarity metric is derived from an overall judgment of the similarity of antigens, antibodies, linkers, and payloads, compared to ADCs in the original dataset. As shown in Table 2, the reconciled mean similarity scores range from 0.4047 to 0.9662, which further reflects the moderate to high dissimilarity between these newly collected ADCs and ADCs in the modelling dataset. Therefore, this external dataset is particularly suitable for testing the generalization ability of the ADCNet model.

Detailed predictive results of the external data set are summarized in Table 2. It is clear that the ADCNet model can accurately forecast the anticancer capacities of 18 out of 19 ADCs, resulting in a prediction accuracy of 94.74%. Accordingly, the independent external testing results not only demonstrate the strong generalization ability of ADCNet in predicting the activity of different types of ADCs, indicating that our model can be used in real-world scenario for ADC drug discovery.

## Table 2. Predictive results of the ADCNet model on an independent external dataset.

| No. | Antibody[1] | Antigen[2] | Payload[3] | Linker | DAR | Similarity score[4] | Bioactivity value | Label | Predictive score |
|---|---|---|---|---|---|---|---|---|---|
| 1 | H1D8 | CD44v5 | MMAE | 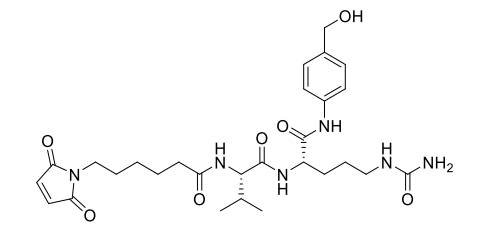 | 3.7 | 0.7867 | $IC_{50}$ = 8.24 μg/mL | 1 | 0.9720 |
| 2 | Cetuximab | EGFR | Gemcitabine | 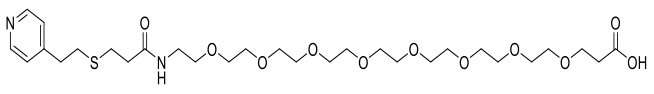 | 6 | 0.4047 | $IC_{50}$ = 10 nM | 1 | 0.9956 |
| 3 | Trastuzumab | HER2 | Gemcitabine | 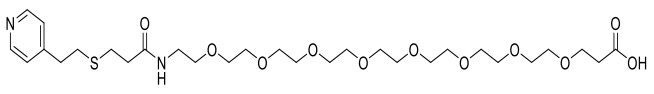 | 6 | 0.4051 | $IC_{50}$ = 150 nM | 0 | 0.9907 |
| 4 | Sacituzumab | TROP2 | Auristatin E | 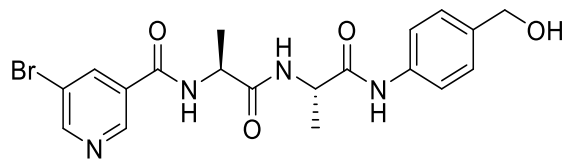 | 6.1 | 0.7240 | $IC_{50}$ = 0.03-0.09 nM | 1 | 0.9839 |
| 5 | Sacituzumab | TROP2 | DM1 | 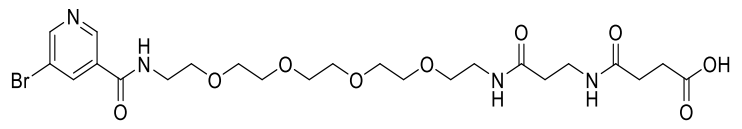 | 6.2 | 0.6541 | $IC_{50}$ = 0.05-0.1 nM | 1 | 0.9991 |
| 6 | Sacituzumab | TROP2 | DXd | 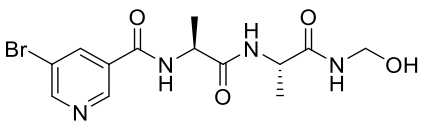 | 7.3 | 0.5757 | $IC_{50}$ = 0.4-0.7 nM | 1 | 0.9817 |
| 7 | Ifinatamab | CD276 | DXd | 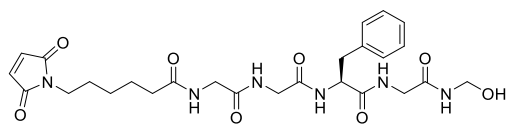 | 8.12 | 0.9662 | $EC_{50}$ = 0.022-3.764 nM | 1 | 0.9889 |
| 8 | Ifinatamab | CD276 | DXd | 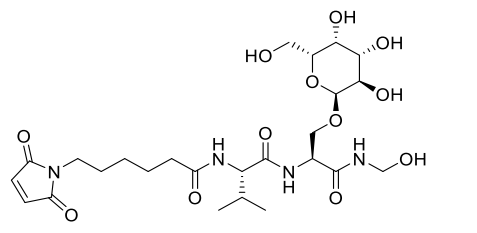 | 8.76 | 0.8152 | $EC_{50}$ = 0.008-3.291 nM | 1 | 0.9326 |
| 9 | Ifinatamab | CD276 | DXd | 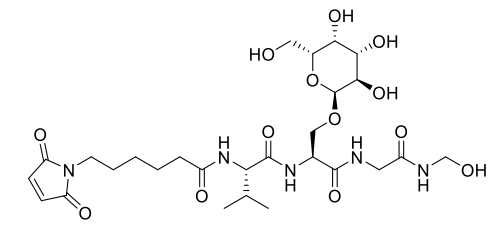 | 7.76 | 0.8019 | $EC_{50}$ = 0.018-2.291 nM | 1 | 0.9648 |
| 10 | Ifinatamab | CD276 | DXd | 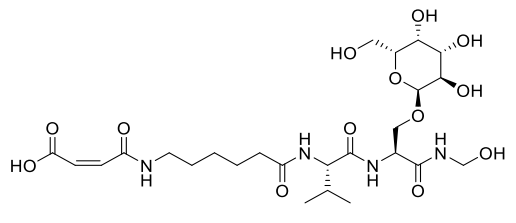 | 7.98 | 0.7733 | $EC_{50}$ = 0.038-1.245 nM | 1 | 0.9946 |
| 11 | Ifinatamab | CD276 | DXd | 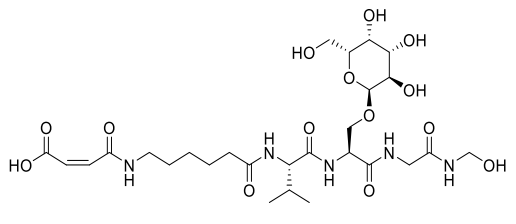 | 8.12 | 0.7616 | $EC_{50}$ = 0.013-0.53 nM | 1 | 0.9950 |
| 12 | Ifinatamab | CD276 | DXd | 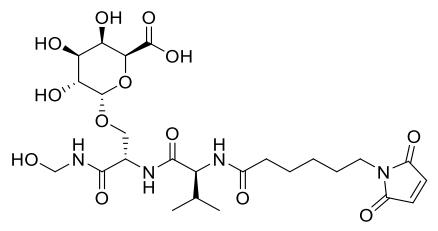 | 8.06 | 0.8107 | $EC_{50}$ = 0.007-0.894 nM | 1 | 0.9847 |
| 13 | Ifinatamab | CD276 | DXd | 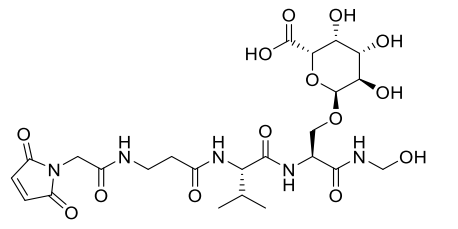 | 8.02 | 0.7271 | $EC_{50}$ = 0.015-1.042 nM | 1 | 0.9946 |
| 14 | Ifinatamab | CD276 | DXd | 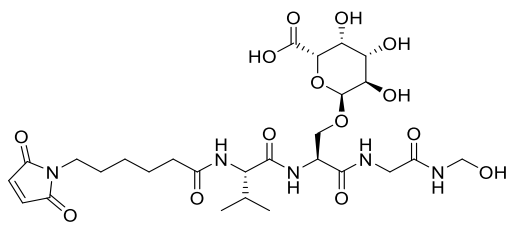 | 8.18 | 0.7976 | $EC_{50}$ = 0.024-1.647 nM | 1 | 0.9910 |
| 15 | Ifinatamab | CD276 | DXd | 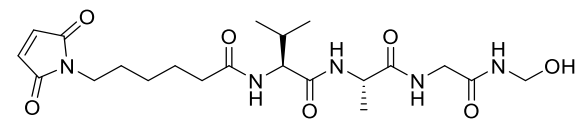 | 8.52 | 0.9125 | $EC_{50}$ = 0.025-1.073 nM | 1 | 0.9855 |
| 16 | Ifinatamab | CD276 | DXd | 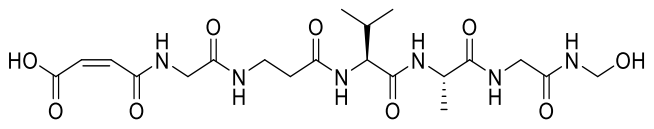 | 8.08 | 0.8544 | $EC_{50}$ = 0.025-1.234 nM | 1 | 0.9958 |
| 17 | Ifinatamab | CD276 | DXd | 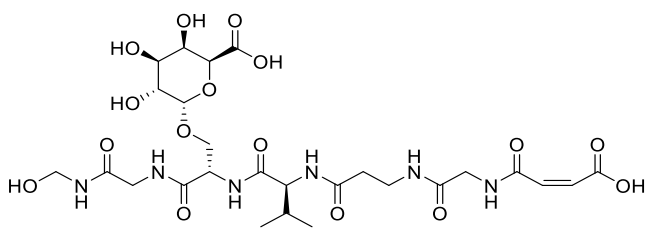 | 7.76 | 0.7273 | $EC_{50}$ = 0.029-1.094 nM | 1 | 0.9958 |
| 18 | Ifinatamab | CD276 | DXd | 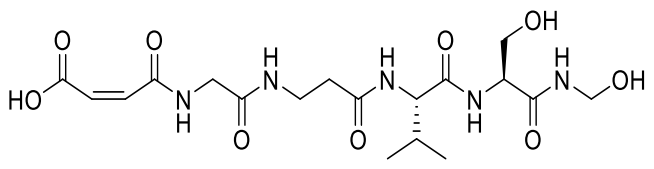 | 7.78 | 0.8074 | $EC_{50}$ = 0.01-0.861 nM | 1 | 0.9958 |
| 19 | Ifinatamab | CD276 | DXd | 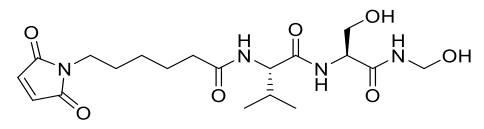 | 8.56 | 0.8914 | $EC_{50}$ = 0.033-0.511 nM | 1 | 0.9616 |

[1]The sequences of antibody are in Supplementary Table S3.

[2] The sequences of antigen are in Supplementary Table S3.

[3]The detailed structures of payloads are in Supplementary Fig. S3.

[4]Similarity score refers to the harmonic mean similarity score, which is used to calculate the average level of similarity between points in a set of data points by taking the reciprocal of the arithmetic mean of the inverse of the similarity score.

### 3.4 Web server construction and usage

To facilitate design and discovery of novel ADC drugs, we developed an online platform called DeepADC based on the optimal ADCNet model for experts and non-experts in the field. DeepADC is freely available at https://ADCNet.idruglab.cn. Users can achieve ADC activity prediction by entering the sequences of antigen and antibody, smiles of linker and payload, and DAR value for a given ADC (Fig. 6). First, user can enter or paste the sequence information of antibody and antigen in the "Antibody Heavy Chain", "Antibody Light Chain", and "Antigen" input boxes (Fig. 6A). Second, users type or paste the smiles of linker and payload of the ADC in the "Linker Iso-SMILES" and "Payload Iso-SMILES" input boxes, respectively, or use the online Molecular Sketcher to draw their molecular structures individually (Fig. 6B). Finally, users enter a DAR value (Fig. 6C) and submit the calculation task. It is worth noting that such DAR value can be determined according to the actual situation, or can refer to the DAR range of the existing antibodies in "Help" function button to determine. Alternatively, users can upload a standard CSV file to perform the activity prediction of multiple ADCs (Fig. 6D). After completing the calculation, users can view the forecast results online (Fig. 6E) or download the forecast results in CSV format.

**Fig 6. The operation of DeepADC.**

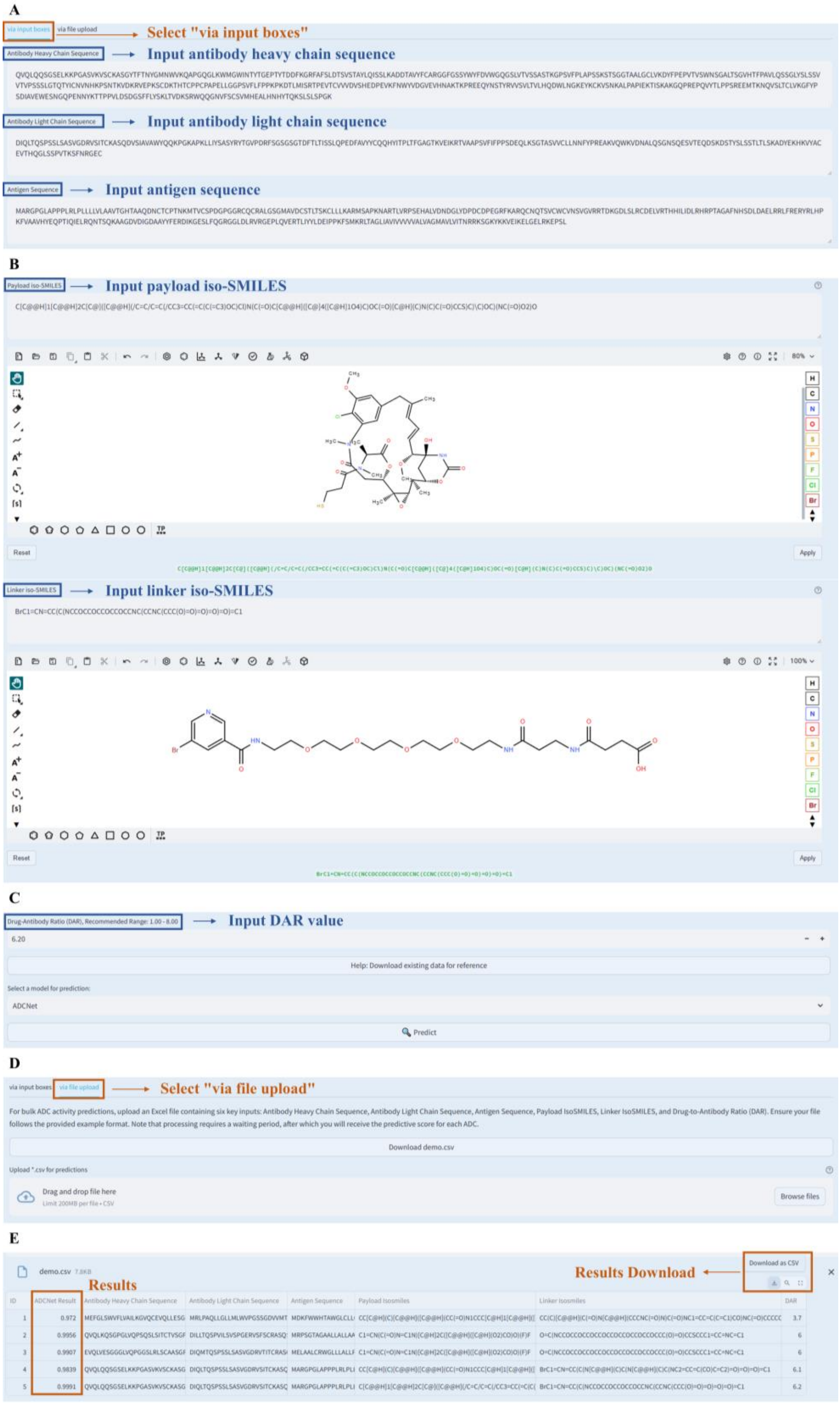

(**A**), (**B**), (**C**), and (**D**) represent the input cases of antibody, antigen, linker and payload, and DAR value, respectively. (**E**) represents the online display or download of the predictive results.

For example, sacituzumab-DM1, a novel ADC constructed by bottom-up aqueous nickel-catalysed cross-coupling[59], was considered as an active ADC due to its high prediction score of 0.9991. Previous studies have demonstrated that sacituzumab-DM1 has strong inhibitory activity against JIMT-1, HCC78, BxPC-3, and MDA-MB-468 cancer cell lines with $IC_{50}$ value of 0.05–0.1 nM, which further confirms the accuracy and availability of DeepADC.

## 4. Conclusion

In this study, we constructed the inaugural DL model called ADCNet, which is able to predict ADC activity. ADCNet creatively incorporates both ESM-2 and FG-BERT LLM models to learn the characteristics of the five components involved in ADCs efficacy. Based on our first carefully-designed ADC benchmark data set, ADCNet achieved the optimal performance on the test set across all evaluation metrics. The 5-fold cross-validation and threshold reclassification experiments successfully confirmed the stability of the model and reduced the influence of human factors, respectively. It is worth mentioning that the ADCNet model achieved the best performance in a series of well-designed ablation experiments, which fully justified the rationally and necessity of the model architecture. Subsequently, external independent testing results also strongly demonstrated the potential of the ADCNet model to handle unknown samples. Finally, to better support the research community in the field, an easy-to-use online platform called DeepADC based on the optimal ADCNet model was constructed to support for the discovery and design of novel ADC drugs.

As the number of other ADC drugs gradually increases, we will build more accurate predictive models and update them to the DeepADC online platform. In addition, we can integrate a variety of biomarkers and drug properties to develop new ADC predictive model in future to improve the accuracy and understanding of models for complex biological systems. In summary, this study opens the door to the rational design of ADC drugs.

**Data availability**

The ADCs data used in the present study are freely available in ADCdb (http://adcdb.idrblab.net/). The remaining data or questions regarding this study are available to the corresponding author upon request (Ling Wang: lingwang@scut.edu.cn). Source data are provided with this paper.

**Code availability**

The source code of ADCNet and associated data preparation scripts are available at github (https://github.com/idruglab/ADCNet). The optimized ADCNet model is also provided.